# Bangla License Plate Recognition Using Convolutional Neural Networks (CNN)

[1]M M Shaifur Rahman, [2]Mst Shamima Nasrin, [1]Moin Mostakim, and [2]Md Zahangir Alom

[1]Department of Computer Science and Engineering, BRAC University, Dhaka, Bangladesh

[2]Department of Electrical and Computer Engineering, University of Dayton, Dayton, OH, USA

Emails: { shaifur.cse, shisshir14}@gmail.com, {nasrinm1, alomm1}@udayton.edu

*Abstract*—In the last few years, the deep learning technique in particular Convolutional Neural Networks (CNNs) is using massively in the field of computer vision and machine learning. This deep learning technique provides state-of-the-art accuracy in different classification, segmentation, and detection tasks on different benchmarks such as MNIST, CIFAR-10, CIFAR-100, Microsoft COCO, and ImageNet. However, there are a lot of research has been conducted for Bangla License plate recognition with traditional machine learning approaches in last decade. None of them are used to deploy a physical system for Bangla License Plate Recognition System (BLPRS) due to their poor recognition accuracy. In this paper, we have implemented CNNs based Bangla license plate recognition system with better accuracy that can be applied for different purposes including roadside assistance, automatic parking lot management system, vehicle license status detection and so on. Along with that, we have also created and released a very first and standard database for BLPRS.

Keywords— *Deep Learning, Convolutional Neural Networks (CNN), Bangla License Plate (BLP), License Plate Recognition (LPR), and Neural Network.*

## I. INTRODUCTION

Automatic License Plate Recognition (LPR) systems are tremendous benefit for traffic, parking, toll management system and cruise control applications [1]. On the other hand, Deep Learning (DL) based approaches have been providing state-of-the-art performance in different recognition tasks [2,3]. The vehicle can be identified with license plate numbers without using any additional information. Nowadays, due to the huge development of transportation sectors and drastic increment of number of vehicles in the road becoming a challenging task to control different types of incident related to the vehicles. The automatic LPR system can ensure better services for traffic monitoring, speeding control, automatic parking lot management system, license number expiration date detection, and many more. When it comes to security monitoring and management of any place or region, LPR systems can be used as a tracking aid to serve as eyes for any security team in any place. In the context of safety and law enforcement, LPR systems play a vital role in border surveillance, physical intrusion and safeguarding [4-7]. The entire system is consisted of different parts including: data acquisition, pre-processing, LPR, and post processing. For data acquisition, the infrared lighting system is used that allow to compute images at any time in day or night. Second, different image enhancement, license plate region detection, individual character extraction, and many more techniques are applied in pre-processing step. After separating characters successfully, the different machine learning approaches including, neural networks, support vector machine (SVM), principle components analysis (PCA), random forest (RF), etc are used to recognize the individual characters. Finally, the recognize license plate is displayed in the display device.

Many kinds of LPR systems are developed using various intelligent computational techniques to obtain accuracy and efficiency [7]. License plate is a principle identification means for any vehicle. Nevertheless, considering fraud circumstances like alteration and replacement of unfit vehicles, LPR systems are correlated with intelligence mechanism for robustness [8]. LPR is one of the very challenging tasks in the filed computer vision, where different techniques are used to deploy a complete system. Recently the DL, Convolutional Neural Networks (CNN) has been showing huge success in different tasks in the field of image processing and computer vision [2,9,33,34] and the CNN has become popular in computer vision community. However, in this implementation, we have used similar convolutional neural network model which has already shown the state-of-the-art recognition accuracy in Bangla handwritten digit recognition [9]. The contribution of this paper can be summarized as follows:

- Very first implementation of Bangla LPR system using CNN method
- We have also created and released a first but standard dataset for Bangla LPR which is available in the link in [35].
- The experiment has been conducted with different number of training and testing samples to verified system performance.
- This work is compared against recently proposed other methods and shows superior performance against equivalent models with the same or fewer number of network parameters.

The paper is organized as follows: Section II discusses related works. The implementation details and CNN model are presented in Section III. Section IV explains the datasets, experiments, and results. The conclusion and future directions are discussed in Section V.

## II. RELATED WORKS

There are different researches have been conducted on LPR system in the last few years and published several papers on it. However, from our knowledge, we did not see any paper on Bangla LPR system with recently developed CNN model. Yu. A. Bolotova, A. A. Druki and V. G. Spitsyn, has proposed a system based on calibrated dual-camera device, which is used to detect license plate of moving vehicles one of them are fixed camera and another one is pan-tilt-zoom camera. This device not only tracks multiple targets but also gets the license plate images with high quality. A CNN is applied to identify the license plate regions and recognizing the alphabets on the license plate [10]. G. R. Goncalves, D. Menotti and W. R. Schwartz proposed on Optical Character Recognition (OCR) based approach which is built with CNN is used for License plate recognition. In this method, the architecture of the CNN is chosen from thousands of random possibilities and its filter weights are set at random and normalized with zero mean and unit norm. The linear SVM based features extraction technique is used with CNN [11]. R. A. Baten, Z. Omair and U. Sikder implement an unique system using "Matra" concept. "Matra" is a feature of Bangla script is used to develop a Bangla License Plate reader. In this proposed model, they have segmented each word as single connected components and later it has recognized through template matching approach for those words [12]. J. Pyo, J. Bang and Y. Jeong have developed a CNN based method which is used to detect Vehicle Recognition. In their proposed architecture first moving car is localized with difference between the frames, the resultant binary image is used to detect the frontal view of a car by symmetry filter, the detected frontal view is used to train and test the CNN [13]. P Wongta, T. Kobchaisawat and T. H Chalidabhongse in their paper, they have described a CNN which is used in multi-oriented Thai-text localization in natural scene images. In this research, text confidence maps are constructed by using specially trained CNN text detector on multi-scaled images. The multi-scaled text confidence maps are merged to produce original input size text confidence map. By using Thai text characteristics, text line hypothesis is used to generate from a merged text confidence map. Finally, the post-processing techniques and Thai text characteristic analysis are performed to acquire text [14]. O. Bulan, V. Kozitsky, P. Ramesh and M. Shreve worked on Automated license plate recognition (ALPR) is used deep localization approach which is actually a strong CNN. Segmentation and Optical Character recognition (OCR) is jointly using a probabilistic inference method based on (Hidden Markov models) HMMs [15]. J. Lwowski and his research team's approach was secure cloud-based deep learning license plate recognition system (LPRS). Their proposed LPRS was used for smart cities which were developed based on deep convolutional neural network. In this proposed model authors used NVIDIA GPUs high performance cloud servers which was pretty much costly [16-18].

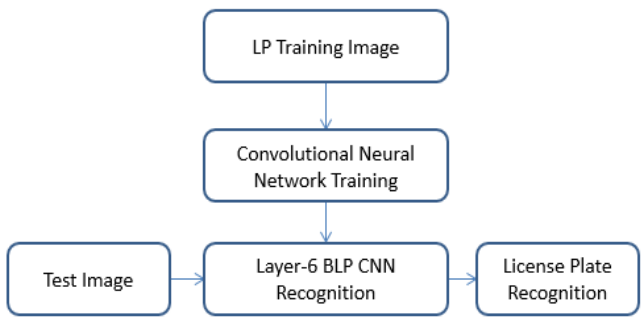

Fig. 1. Overall system implementation diagram where the system is trained with CNN and then the trained model is used for testing phase.

A built-in system was implemented with a GPU in order to recognize the license plate number without detection line. The deep-learning network to recognize the license plate number of the vehicle uses relatively simple AlexNet. Jetson TX1 board was used as it is low cost standalone embedded system specialized device [19]. P. Zemcik and his team has done with a work focused on recognition of license plates in low resolution and low-quality images. The authors present a methodology for collection of real world (non-synthetic) data set of low quality license plate images with ground truth transcriptions. Their approach of the license plate recognition is based on a CNN which holistically processes the whole image, avoiding segmentation of the license plate characters. Evaluation results on multiple datasets show that their method significantly outperforms other free and commercial solutions to license plate recognition on the low-quality sample. To enable further research of low quality license plate recognition, they also make the dataset which is publicly available [20]. However, we have used AlexNet type CNN model for Bangla LPR in this implementation. The implemented Bangla LPR system is evaluated on our standard dataset. The implementation details are represented in the following sections.

### III. IMPLEMENTATION DETAILS

The overall implementation diagram of this proposed system is given in Fig. 1. This system consists of two main parts. First, training section where system is trained with our own dataset for Bangla LRP. After training the system successfully, the testing samples are used to evaluate the testing performance. To release very first and suitable dataset for Bangla License Plate Recognition System (BLPRS), we have considered several steps which are shown in Fig 2.

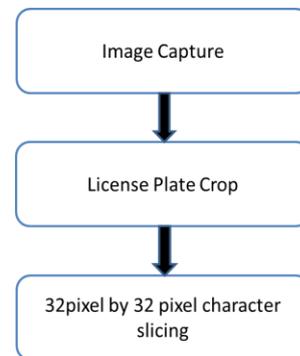

Fig. 2. Steps for data acquisition and dataset preparation.

Data acquisition: we captured nearly three thousand images of license plate. All the images were captured by 12 megapixel camera to ensure better image quality. Images were captured from different locations in Dhaka city which includes indoor parking, outdoor parking, running vehicle from road, vehicle. Images were captured in different condition to make the samples standard. Among the captured images only clearly visible license plate images were taken as primary data to ensure our train dataset more accurate. Images were captured from different angle, time and throughout the year. Fig. 3 (a) and (b) show original images.

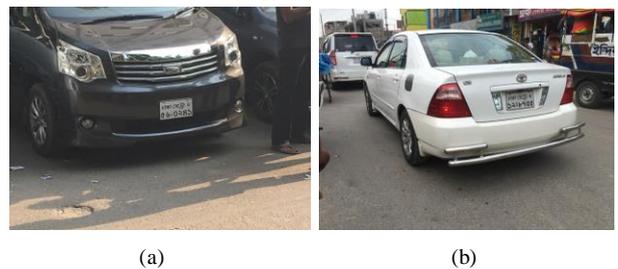

(a)          (b)

Fig. 3. (a) Image from Roadside Parking (b) Running vehicle Image from Road

Data post-processing: license plate was cropped from the main image first; lately digits and characters were sliced in 32pixels by 32 pixels each. That's how all the train set data was prepared. Fig. 4 shows some of our cropped images of license plate from captured images.

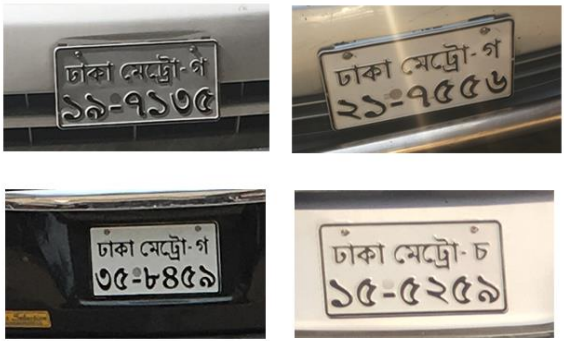

Fig. 4. Manually Cropper License Plate (LP) region from captured samples.

Individual character cropping: at the end, we have cropped individual character from the license plate regions. The Fig. 5, Fig. 6, and Fig. 8 show example database sample. From these sample images, it can be clearly demonstrated that the database contains very challenging samples with different scale, rotation, with different translation, shearing and so on.

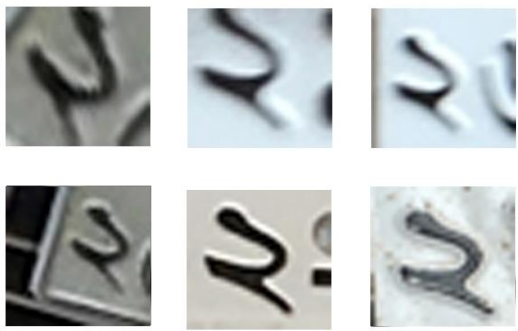

Fig. 5. Sample images for digit 2 from our license plate dataset.

Therefore, this becomes a difficult task to achieve better accuracy for Bangla LPB. However, we have used to CNN model for training and testing and achieved better accuracy for Bangla LPR.

IV. CONVOLUTIONAL NEURAL NETWORK (CNN) FOR BANGLA LPR

The very first version of CNN structure was proposed by Fukushima in 1980 (Fukushima, 1980)[21]. This model has not been vastly applied due to some issue related to the training process. Out of many issues, main issue was not easy to use. LeCun et al. for the very first time has applied a gradient-based learning algorithm with CNN in 1998 and achieved state-of-the-art performance (LeCun et al., 1998a) [22]. This architecture is known as "LeNet". There are further improvement has been made on the LeNet and stated outperformance against traditional methods for different applications. Some of them such as a multi-column CNNs are proposed to recognize digits, alphanumeric, traffic signs, and the other object class (Ciresan & Meier, 2015; Ciresan et al.,2012)[23-24]. They conducted experiment on different publicly available dataset and reported excellent results and surpassed conventional best records on many benchmark databases, including MNIST (LeCun et al., 1998b)[25] handwritten digits database and CIFAR-10 (Krizhevsky & Hinton, 2009)[26,34,35].

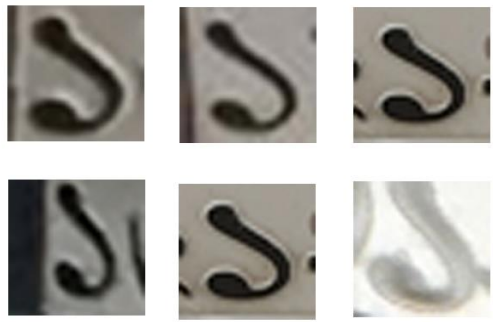

Fig. 6. Sample images for digit 1 from our license plate dataset.

The other advantages of CNN are it has some extra properties: it is more capable to represent a 2D or 3D images with meaningful features which can help to achieve better recognition performance. The max-pooling layer of CNN model is very helpful to deal with shape and scale invariant problems. In addition, due to the sparse connection with tied weights between the layers, CNN requires significantly low number of network parameters compared against a fully connected neural network with similar size. The main point about CNN is trainable with a very popular method which is known as gradient-based learning algorithm. Additionally, there are different modern activation functions are proposed including Rectified Linear Unit (ReLU) which can deal with the diminishing gradient problem. The gradient descent-based method trains the entire model to minimize the errors during training and update the weights accordingly. The CNN produces highly optimized weights during training to ensure better accuracy. In 2014, a deep CNN is used for Hangul (Korean) handwritten character recognition and achieved the best recognition accuracy by Kim & Xie, [27].

Fig. 7 shows an overall CNN architecture which is consisted with two main parts: feature extraction and classification. In the feature extraction part, the layers are used to extract the meaningful features from the inputs samples with different filters where each layer of the network receives the output from its immediate previous layer as its input and passes the current output as input to the next layer. The architecture of CNN consisted of the composed with the combination of three types of layers: convolution, max-pooling, and classification layers. Convolutional layer and max-pooling layers are used in the feature extraction part of the model.

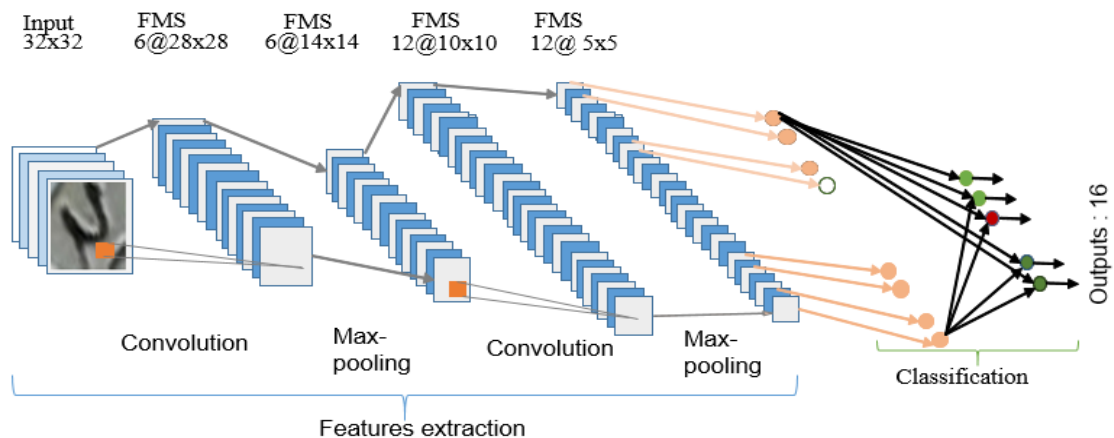

Fig. 7. The overall architecture of the CNN used in this work, which includes an input layer, multiple alternating convolution and max-pooling layers, and one fully connected and classification layers. FMP represent the number of feature maps.

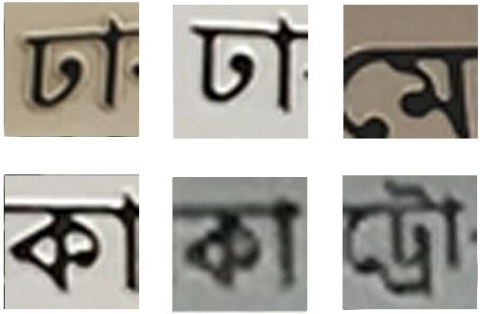

Fig. 8. Sample images for different alphabets from our license plate dataset.

The architecture varies with respect to different implementation and demand. According to the very basic structure of CNN, the even numbered layers are used to perform convolution with different filters and odd numbered layers work for max-pooling operation. The max-pooling operation down samples the outputs samples by the factor of 2. For max-pooling operation, the maximum value is selected from four different pixel values from the input sample. In this implementation, we have used the fully connected layers as a classifier in the classification layer, which has already proved better performance compared to some recent works (Mohamed et al., 2012[28]; Nair & Hinton, 2010) [29]. In the fully connected layer, 300 features have been selected randomly. Finally, the classification layer calculates the class probability and errors have calculated between desire and expected results and backpropagate errors to update weights. In this implementation, we have used the similar model applied for Bangla numeral classification in [9]. In this work, the size of the filters is 5×5 pixels and a feed-forward neural net is used for classification. The sigmoid activation function is used as suggested in most literatures [30,31]. To prevent the overfitting problem, a very efficient technique is used which is called "dropout" (Hinton et al.,2012) [32]. Dropout can help to reduce complexity of the model for several reasons including co-adaptation of neurons, cause a set of neurons are not rely on the presence of another set of neurons. Therefore, it can prevent overfitting the model. However, the main disadvantages of the dropout operations are that it may take more iteration than usual to reach the global minima.

*A. CNN architecture and network parameters*

In this experiment, we have used a model with six different layers including two layers for convolution, two layers for subsampling or pooling, one fully connected layer, and final one classification layer. We have used 6 feature maps in the very first convolutional layer followed by a max-pooling layer. The number of features maps do not change in the pooling layer. In the second convolutional layer, 12 filters are used which produces 12 feature maps from the second convolution layer. After that second max-pooling layer is applied. A fully connected layer is used with 300 neurons which is followed by final classification layer with 16 neurons for 16 different classes. The number of parameters our model is calculated for $32 \times 32$ input image which is shown in Table 1.

Table 1. Number of parameter for different layers of CNN

| Layer | Operation of Layer | Number of feature maps | Size of feature maps | Size of window | Number of parameters |
|---|---|---|---|---|---|
| $C_1$ | Convolution | 6 | 28X28 | 5X5 | 156 |
| $S_1$ | Max-Pooling | 6 | 14X14 | 2X2 | 0 |
| $C_2$ | Convolution | 12 | 10X10 | 5X5 | 1,872 |
| $S_2$ | Max-Pooling | 12 | 5X5 | 2X2 | 0 |
| $F_1$ | Fully Connected | 300 | 1X1 | N/A | 93,600 |
| $F_2$ | Fully connected | 16 | 1X1 | N/A | 4,816 |

For the very first convolutional layers, the size of the output feature maps is $28 \times 28$ and 6 feature maps. The size of the filter mask is $5 \times 5$ for the both convolution layers. The number of network parameters are used to learn is $(5 \times 5 + 1) \times 6 = 156$ and the one is added with filter dimension because of bias. The number of trainable parameters of subsampling layer is 0. However, the size of the feature maps reduced with factor 2 due to 2×2 pooling mask, therefore the size of the outputs of first subsampling layer is $14 \times 14$ with 6 feature maps. The output dimension of the second convolutional layer is (14-5) +1=10, 10×10 with 12 feature maps are used and the second sub sampling filter mask size is 2×2. The learning parameters for second convolution layer are $((5 \times 5+1) \times 6) \times 12 = 1,872$. The zero number of network parameters in second sub-sampling layer. The total number of features per sample from the second sub-sampling layer is 5×5×12 = 300. Since we have used 300 hidden neurons for this layer, therefore, the number of parameters for the first fully connected layer is: $300 \times 12 \times (5 \times 5+1) = 93,600$, whereas the amount of the final layer parameter is: $16 \times (300+1) = 4,816$. Total number of parameters is 1,00,444 for this model.

V. EXPERIMENT AND RESULT

To evaluate the performance of the CNN models for Bangla LPR, we have tested on our own dataset. For this implementation, we have used the MATLAB version R2015a, MATLAB based deep learning framework were used to implement CNN which is run on a single machine with 8 of RAM. We have conducted experiment with our own dataset which contains 1750 number of training samples and 350 number of testing samples. We did not apply any data augmentation or data processing steps for preparing this dataset in pre-processing step. According to the workflow of our system which is illustrated in Fig. 1. First the CNN was trained with prepared dataset of sixteen different classes which consists of zero to nine and some Bangla character which have observed very frequently on Bangla license plate. We prepare total one thousand and nine hundred dataset where each class has more than one hundred numbers of samples. Bangla License Plate

Recognition System (BLPRS) is trained with CNN which is evaluated with different number of samples in dataset. We divide the dataset into train set and test set in different fold like 70% data is used to train 30% to test then again 80% to train and 20% to test 90% to train and 10% of data to test to verify the impact of number of samples on testing performance.

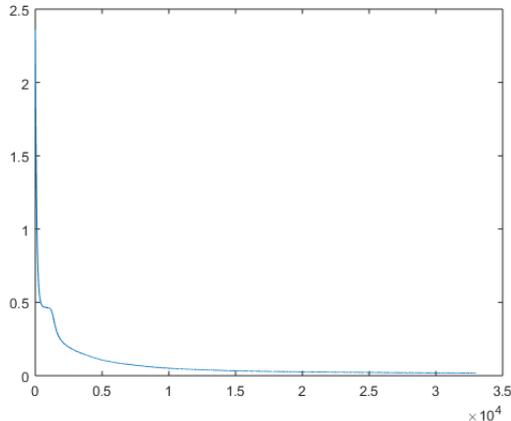

Fig. 9. Tradeoff between number of Iteration versus errors for 1000 epochs for training.

*A. Results and discussion*

During the experiment, the entire experiment has been run for several times for different combination of training and testing samples. The training accuracy versus the number for 1000 epochs (which is the highest number of epochs) is shown in Fig. 9. From different trials with different number of epochs and different split ratio, it can be clearly observed that if we increase the number of epochs then the model converged properly with the lowest errors. From the Fig. 9, it can be clearly observed that the errors of the proposed system decrease with respect to the number of epochs. We have conducted experiment for the different number of epochs including 100, 500, and 1000 epochs respectively. In each case, the model has been tested with the samples from the testing samples. We have achieved different performance for different trained models. The summary of the different experiment with testing accuracy is shown in Table 2. As deep learning is a data driven techniques, here we have only uses 1750 number of samples for training the whole system. Therefore, by increasing the number of epoch or iterations we have tried to ensure better learning during training.

Finally, we have tested our proposed techniques with all the different trained models and achieved different testing accuracy with respect to the number of inputs sample in the testing phase which is shown in Fig. 10. This figure shows the testing accuracy for different trials with different number of samples where column represents the trail numbers 1, 2, 3, 4 where the model has trained with 1300, 1500, 1650, and 1750 number of training samples respectively. We have achieved 88.67% testing accuracy for highest number of training samples which is the highest testing accuracy in this implementation. However, with the other training samples, we have achieved 81.09%, 82.28% testing accuracy respectively. Therefore, it can be clearly summarized that the number of training samples has huge impact on the performance of deep learning approaches.

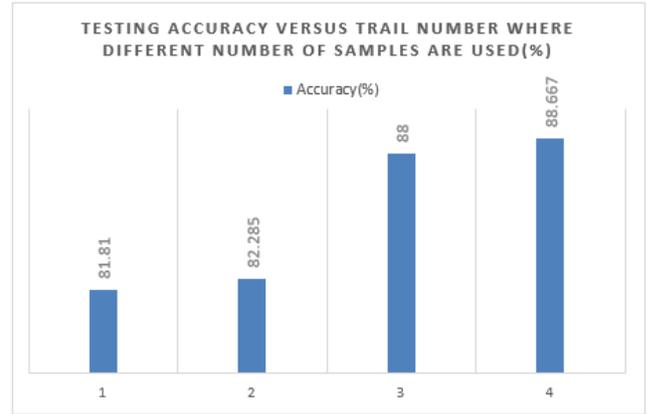

Fig. 10. Testing accuracy for different number of input samples

After observing the Table 2, we can be concluded that large number of samples and epoch provides better accuracy for training and testing phase. However, larger number of epochs take longer time to produce output.

Table 2. Different experimental results with number of samples, number of epochs, total computational time, time per epoch and testing accuracy.

| Train Sample | Epoch | Computational Time(Sec) | Average time/Epoch | Accuracy (%) |
|---|---|---|---|---|
| 1300 | 100 | 217.62 | 2.1762 | 70 |
| 1300 | 500 | 1185 | 2.37 | 81.8181 |
| 1300 | 1000 | 2208 | 2.208 | 81.0909 |
| 1500 | 100 | 264.22 | 2.6422 | 73.428 |
| 1500 | 500 | 1262.28 | 2.52456 | 82.285 |
| 1500 | 1000 | 2641.15 | 2.64115 | 82.285 |
| 1650 | 100 | 283.5 | 2.835 | 79.33333 |
| 1650 | 500 | 1483.6 | 2.9672 | 88.66667 |
| 1650 | 1000 | 2762.5 | 2.7625 | 88 |
| 1750 | 100 | 287.4 | 2.874 | 82 |
| 1750 | 500 | 1783.5 | 3.567 | 86 |
| 1750 | 1000 | 2890 | 2.89 | 88 |

*B. Limitations*

The limitation that was faced for training was due to having smaller memory and computational power. A general-purpose computer with 8GB of RAM and core i3 3.2GHz CPU was used to train and test the system. As for our training and simultaneous testing, the memory and computational power requirement was high but due to lack of resources the testing process takes more time. For using any other proponent or using deeper CNN model, the more memory is required. If we would able to train and test the model on GPU, the time required would be reduced by the factor of at least 10. A large CNN model on GPU system surely will improve overall performance for Bangla LPR.

VI. CONCLUSION AND DISCUSSION

A convolutional neural network is a cornerstone for any recognition task nowadays ranging from classification, detection to segmentation tasks. In addition, convolutional neural networks are better feature-extractors than any fully connected layer consisting counterparts. Our idea is to implement a system for Bangla License Plate Recognition

(BLPRS) with CNN which is evaluated with a set of experiment. The experimental results demonstrate that CNN approach is capable to obtain better performance. In this implementation, we have achieved around 89% testing accuracy for Bangla License Plate Recognition System (BLPRS). Another contribution of this work is that we have created and released a new benchmark dataset for BLPRS which will be helpful for implement and evaluated BLPRS further. In the future, we would like to use bigger dataset for training our model on GPU to achieve further better performance. Moreover, the implemented Convolutional Neural Network (CNN) consists of 6 layers including a fully connected layer and with only 6 and 12 feature maps in two convolution layers respectively. It can be concluded that the performance of this proposed license place recognition system will be increased significantly with higher number of features maps and more layers which is another future direction of this work.